\documentclass[conference]{IEEEtran}
\IEEEoverridecommandlockouts

\usepackage{cite}
\usepackage{amsmath,amssymb,amsfonts}
\usepackage{algorithmic}
\usepackage{graphicx}
\usepackage{textcomp}
\usepackage{xcolor}
\usepackage{subcaption}                  
\usepackage{caption}                     
\usepackage{hyperref}  
\usepackage{array} 
\usepackage{titlesec}
\usepackage{afterpage}
\usepackage{lineno}
\usepackage{array} 
\usepackage{adjustbox}
\usepackage{multirow} 
\usepackage{bigstrut} 
\usepackage{lineno}

\newcolumntype{P}[1]{>{\centering\arraybackslash}p{#1}}

\def\BibTeX{{\rm B\kern-.05em{\sc i\kern-.025em b}\kern-.08em
    T\kern-.1667em\lower.7ex\hbox{E}\kern-.125emX}}
\begin{document}

\title{Using Shape Metrics to Describe 2D Data Points
}

\author{\IEEEauthorblockN{ William Franz Lamberti}
\IEEEauthorblockA{\textit{Center for Public Health Genomics} \\
\textit{Department of Biomedical Engineering}\\
\textit{School of Data Science}\\
\textit{University of Virginia}\\
Charlottesville, VA, United States \\
william.f.lamberti@virginia.edu}
}

\maketitle
\IEEEpubidadjcol

\begin{abstract}
Traditional machine learning (ML) algorithms, such as multiple regression, require human analysts to make decisions on how to treat the data.  These decisions can make the model building process subjective and difficult to replicate for those who did not build the model.  Deep learning approaches benefit by allowing the model to learn what features are important once the human analyst builds the architecture.  Thus, a method for automating certain human decisions for traditional ML modeling would help to improve the reproducibility and remove subjective aspects of the model building process.  To that end, we propose to use shape metrics to describe 2D data to help make analyses more explainable and interpretable.  The proposed approach provides a foundation to help automate various aspects of model building in an interpretable and explainable fashion.  This is particularly important in applications in the medical community where the `right to explainability' is crucial.  We provide various simulated data sets ranging from probability distributions, functions, and model quality control checks (such as QQ-Plots and residual analyses from ordinary least squares) to showcase the breadth of this approach.
\end{abstract}


\section{Introduction}

In the age of big data, 2D representations of multivariate data are still commonplace and a requirement for many experiments\cite{mcinnes_umap_2020, mateo_visualizing_2019, takei_integrated_2021, alexandrov_spatial_2020, hotelling_analysis_1933}.  For example, analysts continue to simplify results to 2D representations by analyzing various underlying properties and implicit assumptions.  However, many of these analyses are subjective and not numeric.  For instance, analyzing the residuals of multiple regression is vital to ensure that the assumptions of the model are met\cite{mendenhall_second_2011, draper_applied_1998}.  However, these checks are usually checked by human analysts and are not verified using numeric measures.  There are many other aspects of multiple regression that require human decisions to be made such as the inclusion of interaction terms.  These human decisions can be a downside when compared to deep learning approaches since deep learning models are able to learn what features are important for a given analysis without the need for a human analyst\cite{gu_recent_2018}.  Having these human elements makes experiments using these approaches less explainable and interpretable for those not involved in the experiment.  

Ensuring that a computational analysis is as explainable and interpretable as possible is key for explainable artificial intelligence (XAI) applications.  XAI is used in various fields such as medicine and national security\cite{lamberti_william_franz_overview_2022}.  For example, the European Union has passed laws ensuring a patient's ``right to explainability''\cite{european_union_regulation_2016}.  In short, this law states that if a computational model is used to help make a diagnosis, the aspects of the computational model must be able to be described in layman's terms.  However, XAI is not limited to the intersection of science and policy.  XAI is key for making scientific inferences since scientists need to understand how AI models use features to better understand the scientific phenomena.  Thus, there is a fundamental need to make computational analyses as intuitive and clear as possible.  A shape metric like area is a prime example of a metric which has a clear definition and is a concept that is understandable to the average person\cite{lamberti_william_franz_overview_2022}.  Thus,  we desire to use explainable and interpretable shape metrics to describe data to help quantify the shapes of 2D data.  This paper posits that all data that resides in 2D feature spaces can be represented as images.  These images can be analyzed by extracting various useful shape metrics.

The work that provides the foundation for the idea of analyzing 2D data using shape metrics is eigenvale decomposition.  Eigenvalues correspond to the relative length of the axes of the data\cite{lamberti_william_franz_overview_2022}.  For example, if we observe a 2D scatterplot, the major and minor axis lengths will be captured by the first and second eigenvalues, respectively \cite{lamberti_william_franz_overview_2022}.  Thus, eigenvalue decomposition is the first idea that measures the shape of data.  However, this idea has not been significantly expanded.  To that end, we are providing foundational evidence and an approach to use a variety of shape metrics to describe data in 2D feature spaces.   

\begin{figure*}[h!]
\centering 
\includegraphics[width=0.98\linewidth]{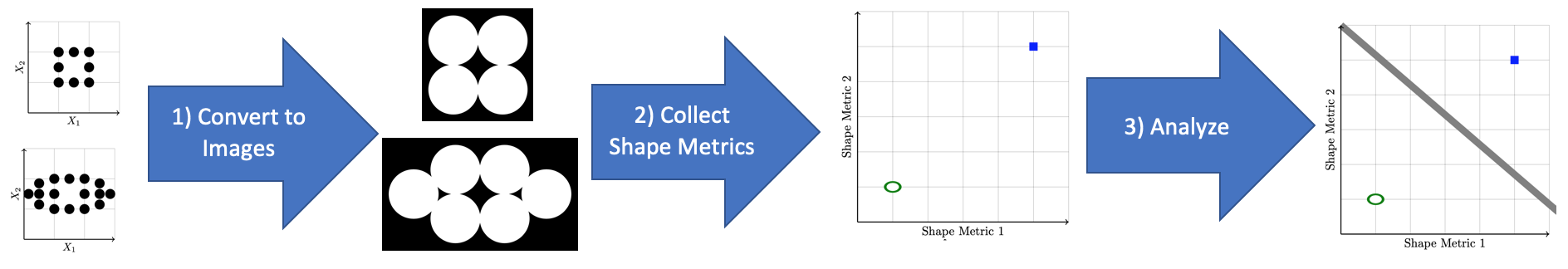}
\caption{Overview of analyzing 2D data as images using shape metrics.  Once the 2D data is collected, step 1) converts the data to binary images.  Step 2) collects various shape metrics of interest to describe the images. Step 3) analyzes the shape metrics.  In this example, the differently shaped data points are classified.  However, other analyses could be performed.  }
\label{fig: overview}
\end{figure*}

We provide a new manner to analyze 2D data as images.  We first convert the data in 2D space to 2D images of the shapes.  We then collect shape metrics from the images.  We lastly analyze the images for a given analysis.  By quantifying 2D data using tangible shape metrics, we make analyses of 2D data more explainable and interpretable.  An overview of our contribution is provided in Figure \ref{fig: overview}.  The code for our experiments are provided at our GitHub link: \url{https://github.com/billyl320/2d_shape_points}.

\section{Methods and Materials}

There are various simulated scenerios we will analyze: the discrimination between different Normal distributions, the detection of outliers in QQ-Plots, the discrimination of different 2D functions, and the analysis of multiple regression or ordinary least squares (OLS) residual analysis of variance.  

\textbf{Normal Distributions:} The discrimination of different Normal distributions will help to provide a foundation by which we are able to understand how shape metrics are useful on conceptually concrete examples.  We will denote Normal or Gaussian distributions with a mean $\mu$ and variance $\sigma^2$ as $N(\mu, \sigma^2)$.  The first experiment will have two simulated Normal distributions with a common variance, but different means.  This experiment will show that the means of the distribution are not important for discriminating the distributions.  The following two experiments will have pairs of Normals with the same mean but differing variances.  These two experiments provide evidence that the shape metrics are useful for discriminating Normals with differing variances.  

\begin{figure}[h!]
\centering 
\begin{subfigure}[t]{.24\textwidth}
\centering 
\includegraphics[width=0.90\linewidth]{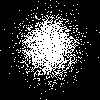}
\caption{1 Gaussian}
\label{fig: mix1}
\end{subfigure}
\centering 
\begin{subfigure}[t]{.24\textwidth}
\centering 
\includegraphics[width=0.90\linewidth]{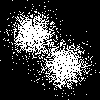}
\caption{2 Gaussians}
\label{fig: mix2}
\end{subfigure}
\begin{subfigure}[t]{.24\textwidth}
\centering 
\includegraphics[width=0.90\linewidth]{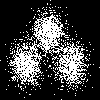}
\caption{3 Gaussians}
\label{fig: mix3}
\end{subfigure}
\begin{subfigure}[t]{.24\textwidth}
\centering 
\includegraphics[width=0.90\linewidth]{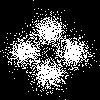}
\caption{4 Gaussians}
\label{fig: mix4}
\end{subfigure}
\caption{Examples of mixtures of Gaussians.  }
\label{fig: mix_exs}
\end{figure}

The followup to these experiments uses various mixtures of Gaussian distributions.  This will show that different mixtures of Normals can be discriminated using shape metrics.  Examples are provided in Figure \ref{fig: mix_exs}.  Thus, this set of four experiments using a variety of different Normal distributions shows that the variance and the number of Gaussians mixtures present are key for providing differently shaped data.

\begin{figure}[h!]
\centering 
\begin{subfigure}[t]{.24\textwidth}
\centering 
\includegraphics[width=0.90\linewidth]{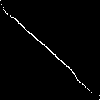}
\caption{No outliers}
\label{fig: q1}
\end{subfigure}
\centering 
\begin{subfigure}[t]{.24\textwidth}
\centering 
\includegraphics[width=0.90\linewidth]{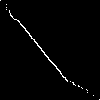}
\caption{Minor outliers}
\label{fig: q2}
\end{subfigure}
\begin{subfigure}[t]{.24\textwidth}
\centering 
\includegraphics[width=0.90\linewidth]{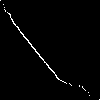}
\caption{Medium outliers}
\label{fig: q3}
\end{subfigure}
\begin{subfigure}[t]{.24\textwidth}
\centering 
\includegraphics[width=0.90\linewidth]{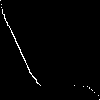}
\caption{Major outliers}
\label{fig: q4}
\end{subfigure}
\caption{QQ-Plot examples with different levels of  outliers.  }
\label{fig: qqs}
\end{figure}

\textbf{QQ-Plots and Outlier Detection:} The second experiment aims to show that shape metrics can be used to classify QQ-Plots that have outliers.  QQ-Plots are a visualization technique used to ensure that data follows a particular distribution\cite{wilk_probability_1968}.  For OLS, we want the residuals to follow a Normal distribution \cite{bhattacharyya_statistical_1977, mendenhall_second_2011, draper_applied_1998}. We simulated various QQ-Plots with no outliers (1000 random $N(0,1)$), minor outliers (990 random $N(0,1)$ and 10 random $N(3,1)$), medium outliers (990 random $N(0,1)$ and 10 random $N(5,1)$), and major outliers (990 random $N(0,1)$ and 10 random $N(10,1)$) to showcase the ability of our approach to identify outliers.  Examples of these QQ-Plots as images are provided in Figures \ref{fig: q1} - \ref{fig: q4}, respectively.  

\begin{figure}[h!]
\centering 
\begin{subfigure}[t]{.24\textwidth}
\centering 
\includegraphics[width=0.90\linewidth]{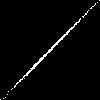}
\caption{Linear}
\label{fig: lin}
\end{subfigure}
\centering 
\begin{subfigure}[t]{.24\textwidth}
\centering 
\includegraphics[width=0.90\linewidth]{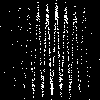}
\caption{Sine}
\label{fig: sin}
\end{subfigure}
\centering 
\begin{subfigure}[t]{.24\textwidth}
\centering 
\includegraphics[width=0.90\linewidth]{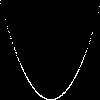}
\caption{Parabola}
\label{fig: poly1}
\end{subfigure}
\centering 
\begin{subfigure}[t]{.24\textwidth}
\centering 
\includegraphics[width=0.90\linewidth]{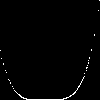}
\caption{Poly}
\label{fig: poly2}
\end{subfigure}
\caption{Examples of the various simulated functional models resulting images.}
\label{fig: mods}
\end{figure}

\textbf{Functions:} The third experiment aims to show that different shaped 2D functions can be discriminated using shape metrics.  Each function had 1000 random simulated observations per resulting image.  The first function was\begin{equation}\label{eq: ols}
    Y = 3X +\epsilon    
\end{equation}
\noindent where $X\sim N(0, 100)$ and $\epsilon\sim N(0,1)$.  This is referred to as the ``Linear'' function.  Note that $\sim$ represents ``distributed as''.  The second function was  \begin{equation}\label{eq: sin}
    Y = 4\sin{X} +\epsilon    
\end{equation}
\noindent where $X\sim N(0, 100)$ and $\epsilon\sim N(0,0.25)$.  This is referred to as the ``Sine'' model.  The third function was\begin{equation}\label{eq: sqrd}
    Y = X^2 +\epsilon    
\end{equation}
\noindent where $X\sim N(0, 100)$ and $\epsilon\sim N(0,1)$.  This is referred to as the ``Parabola'' function.  The fourth function was\begin{equation}\label{eq: poly}
    Y = X^4 + 10X^3 - 7X^2 +\epsilon    
\end{equation}
\noindent where $X\sim N(0, 100)$ and $\epsilon\sim N(0,1)$.  The is referred to as the ``polynomial'' or ``Poly.'' model.  Examples of these functions are provided in Figures \ref{fig: lin} - \ref{fig: poly2}, respectively.  

\begin{figure}[h!]
\centering 
\begin{subfigure}[t]{.24\textwidth}
\centering 
\includegraphics[width=0.90\linewidth]{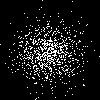}
\caption{Random normal scatter}
\label{fig: ran}
\end{subfigure}
\centering 
\begin{subfigure}[t]{.24\textwidth}
\centering 
\includegraphics[width=0.90\linewidth]{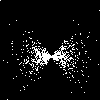}
\caption{Multiplicative scatter}
\label{fig: bow}
\end{subfigure}
\caption{Examples of a subset of the OLS residual simulated resulting images.   }
\label{fig: residuals}
\end{figure}

\textbf{OLS Residual Analysis:} The fourth and last experiment corresponds to evaluating different types of variance plots for multiple regression.  The plots should have random scatter and not display any obvious patterns.  An example of this is provided in Figure \ref{fig: ran}.  Cone-like shape could indicates an increase in variance as the response increases.  This is common for Poisson phenomena.  Another concerning pattern looks like an almond or an eye.  This occurs when the response is a proportion from a Binomial random variable (shorted to ``Binom'').  The last error discussed here is multiplicative errors (shorted to ``Multi.''), where the plot will look like a bowtie.  An example of multiplicative errors is provided in Figure \ref{fig: bow}.  Once these patterns are identified, transformations can be applied to the response to correct for these errors\cite{mendenhall_second_2011}.  Automating these transformations would help to make multiple regression more standardized and reproducible.  

\textbf{Converting Data to Images:} The crucial step is converting the raw data into images.  This is done by converting the images using 2D histograms and then accepting all positive signals.  We will describe this more precisely using image operator notation\cite{kinser_image_2018}.  Thus, 2D raw data are converted into 2D images using  \begin{equation}
    \textbf{b}[\vec{x}] = \Gamma_{>0}\mathbb{H}_{2, 2} X,  
\end{equation}
\noindent  where $X$ is the input data, $\mathbb{H}_{2, 2}$ converts the data into a 2D histogram\cite{noauthor_numpyhistogram2d_nodate}, and $\Gamma_{>0}$ is the threshold image operator.  From here, we would collect various shape metrics of interest.  

\textbf{Shape Metrics:} Due to space limitations, we do not provide the exact manner in which these shape metrics are collected.  However, we will provide a brief description of the shape metrics used in this analysis.  Much of the description of these metrics is borrowed from Lamberti\cite{lamberti_classification_2020}.  Lamberti's extended descriptions on the shape metrics used in this analysis can be found at various sources\cite{lamberti_algorithms_2020, lamberti_classification_2020, lamberti_william_franz_overview_2022}.  The first metrics were the shape proportions (SP) and encircled image-histograms (EI), which were collected from the shape proportion and encircled image-histogram (SPEI) algorithm \cite{lamberti_algorithms_2020}.  The EI is the black and white pixel counts of the shape after the shape is placed in the minimum encompassing circle and then the minimum encompassing square \cite{lamberti_algorithms_2020}.  In other words, this is the area and the surrounding area of the shape.  The SP value is the proportion of the area of the shape relative to the sum of the EI.  The other shape metrics collected that were used in the model were the eigenvalues of the shapes, eccentricity \cite{kinser_image_2018}, and circularity \cite{kinser_image_2018, rosenfeld_compact_1974}.  The eigenvalues measure the major and minor axes of the shape.  Eccentricity is the ratio of the major axis over the minor one.  These are calculated using the $1^{\text{st}}$ and $2^{\text{nd}}$ eigenvalues of the shapes, respectively \cite{kinser_image_2018}.  Circularity measures how circular a given shape is.  This results in a total of 7 total metrics used during our analyses.  The used metrics are summarized in Table \ref{tab:metrics}.

\begin{table}[h!]
\centering
\caption{ Table provides the metrics used in this analysis on a given image, $i$.  The first column is the $q^{\text{th}}$ metric, where $q \in \{1, 2, ..., 7\}$.  The last column provides the number of times the classification trees used each metric.  These variables make our models interpretable and explainable\cite{lamberti_william_franz_overview_2022}.  }
\begin{tabular}{c|cc}
  \hline
$\vec{m}_{q,i}$ & Metric & Counts \\ 
  \hline
  1      & White EI & 6\\ 
  2      & Black EI & 0 \\ \\
  3      & SP value & 3 \\ 
  4      & Eccentricity & 4  \\  \\
  5      & $1^{\text{st}}$ Eigenvalue & 2 \\ 
  6      & $2^{\text{nd}}$ Eigenvalue & 0 \\  \\
  7      & Circularity & 1 \\
   \hline
\end{tabular}
\label{tab:metrics}
\end{table}

\textbf{Modeling:} Once we collected the 7 metrics, collected 100 simulated cases (which results in 100 images) for each class for a given experiment.  We used a classification tree to discriminate between the different classes\cite{therneau_rpart_2019}.  We used 80\% of the data as training and 20\% as validation.  We used stratified random sampling to ensure that the proportions were evenly split between the different classes.  On the training data, we used 5-fold cross-validation to select the complexity parameter\cite{james_introduction_2013, takei_integrated_2021}.  

\section{Results}\label{sec:res}
A summary of the results are provided in Table \ref{tab:rslts}.  Each set of experiments is analyzed in more detail in the following sections.  As a whole, these experiments show that raw 2D data can be analyzed as images using their shape metrics. 

\begin{table}[h!]
    \centering
    \caption{Validation accuracy measures and confidence intervals (CIs) of classification trees for different experiments.  The first 4 rows correspond to the comparison of multivariate Normal distributions under different conditions.  The fifth row corresponds to the QQ-Plot experiment.  The sixth row corresponds to the Function experiment.  The seventh row corresponds to the OLS residual analysis experiment.  Note that $\mu = \begin{bmatrix}
0 \\
0 
\end{bmatrix}$ and $\Sigma = \begin{bmatrix}
1 & 0 \\
0 & 1 
\end{bmatrix}$.  }
    \label{tab:rslts}
    \begin{tabular}{c|cc}
        Experiment & Accuracy & Accuracy 95\% CI \\ \hline \\
        $N_1(\begin{bmatrix}
0 \\
0 
\end{bmatrix}, \Sigma), N_2(\begin{bmatrix}
10 \\
10 
\end{bmatrix},\Sigma)$  & 0.525 & (0.3613, 0.6849) \\ \hline \\
        $N_1(\mu, \Sigma), N_2(\mu,\begin{bmatrix}
1 & 0.9 \\
0.9 & 1 
\end{bmatrix})$  & 1.00 & (0.9119, 1.00)\\ \hline \\
        $N_1(\mu, \Sigma), N_2(\mu,\begin{bmatrix}
0.001 & 0 \\
0 & 0.001 
\end{bmatrix})$  & 1.00 & (0.9119, 1.00)\\ \hline \\
        4 Gaussian Mixtures  & 0.975 & (0.9126, 0.997) \\ \hline \hline \\
        QQ-Plots and Outlier Detection & 0.95 &(0.8769, 0.9862)\\ \hline \hline \\
        Functions & 0.9375 &(0.8601, 0.9794)\\ \hline \hline \\
        OLS Residual Analysis & 0.9375 &(0.8601, 0.9794)
    \end{tabular}
\end{table}

\textbf{Normal Distributions}: This analysis shows that shape metrics are useful for only discriminating the shape of data, not the location of the data.  The first experiment shows where shape metrics will be unhelpful as only the means differ between the distributions.  The shape of the distribution is primarily described by the variance of the distribution.  

\textbf{QQ-Plots and Outlier Detection}: Despite having varying levels of outliers, we were able to discriminate between different levels of outliers on our QQ-Plots.  This shows that the shape metrics are useful in quantifying different levels of outliers present in 2D data.  This would help to quantify and detect outliers automatically. Thus, we have evidence that shape metrics can be used to detect the presence of outliers automatically.

\textbf{Functions}: Shape metrics are useful for classifying different kinds of 2D functions from one another.  This is useful for evaluating a variety of different functional shapes.  This can be used to help guide analysts on the kind of model to utilize for a given analysis.  This experiment provides evidence for automatic function determination.  

\textbf{OLS Residual Analysis}: Shape metrics are useful for determining the kind of variance pattern observed in multiple regression residual plots.  This helps to remove the more subjective aspects of model evaluation and help to provide a standard and reproducible process for modeling choices.  This provides evidence for automated OLS residual analysis using shape metrics.   

\textbf{Useful Features}: Table \ref{tab:metrics} shows the number of times each of the metrics collected were used across all of the experiments in the classification trees.  The White EI (or area) was used the most out of all of the chosen metrics.  Eccentricity was the second most used metric.  Thus, White EI and Eccentricity should be used in future analyses and applications at a minimum.  

\section{Conclusions}  Shape metrics are useful for describing a variety of different 2D data scenarios.  This provides strong evidence that all 2D raw data are images and can be analyzed as such.  Thus, we can standardize traditional analyses like multiple regression to help automate modeling decisions and make them more explainable and interpretable to those not involved in the analysis.  This paper provides a strong foundation for analysts to analyze 2D data as images using shape metrics.  

\section{Acknowledgements}

UVA Engineering Graduate Writing Lab Peer Review Group provided valuable feedback during initial drafts of this paper.  We would also like to thank the Zang Lab for Computational Biology at the University of Virginia for their support. 

\clearpage

\bibliographystyle{IEEEtran}
\bibliography{Zotero.bib}

\end{document}